\documentclass{article}

\usepackage{PRIMEarxiv}

\usepackage[utf8]{inputenc} 
\usepackage[T1]{fontenc}    
\usepackage{hyperref}       
\usepackage{url}            
\usepackage{booktabs}       
\usepackage{amsfonts}       
\usepackage{nicefrac}       
\usepackage{microtype}      
\usepackage{lipsum}
\usepackage{fancyhdr}       
\usepackage{graphicx}       
\graphicspath{{media/}}     

\title{Comparison of Object Detection Algorithms for Street-level Objects
}

\author{
 Martinus Grady Naftali \\
  School of Computer Science\\
  Bina Nusantara University\\
  Jakarta, Indonesia \\
  \texttt{martinus.naftali@binus.ac.id} \\
   \And
 Jason Sebastian Sulistyawan \\
  School of Computer Science\\
  Bina Nusantara University\\
  Jakarta, Indonesia \\
  \texttt{jason.sulistyawan@binus.ac.id} \\
  \And
 Kelvin Julian \\
  School of Computer Science\\
  Bina Nusantara University\\
  Jakarta, Indonesia \\
  \texttt{kelvin.julian@binus.ac.id} \\
}

\begin{document}
\maketitle

\begin{abstract}
Object detection for street-level objects can be applied to various use cases, from car and traffic detection to the self-driving car system. Therefore, finding the best object detection algorithm is essential to apply it effectively. Many object detection algorithms have been released, and many have compared object detection algorithms, but few have compared the latest algorithms, such as YOLOv5, primarily which focus on street-level objects. This paper compares various one-stage detector algorithms; SSD MobileNetv2 FPN-lite 320x320, YOLOv3, YOLOv4, YOLOv5l, and YOLOv5s for street-level object detection within real-time images. The experiment utilizes a modified Udacity Self Driving Car Dataset with 3,169 images. Dataset is split into train, validation, and test; Then, it is preprocessed and augmented using rescaling, hue shifting, and noise. Each algorithm is then trained and evaluated. Based on the experiments, the algorithms have produced decent results according to the inference time and the values of their precision, recall, F1-Score, and Mean Average Precision (mAP). The results also shows that YOLOv5l outperforms the other algorithms in terms of accuracy with a mAP@.5 of 0.593, MobileNetv2 FPN-lite has the fastest inference time among the others with only 3.20ms inference time. It is also found that YOLOv5s is the most efficient, with it having a YOLOv5l accuracy and a speed almost as quick as the MobileNetv2 FPN-lite. This shows that various algorithm are suitable for street-level object detection and viable enough to be used in self-driving car.
\end{abstract}


\section{Introduction}
Computer Vision and Artificial Intelligence have a subset known as Object Detection. Where its purpose is to develop a machine learning model that is capable of detecting certain objects within an image \cite{b1, b2}. Object detection itself can be seen as one of the fundamental concepts used in many real-world machine learning applications such as image recognition, face detection and many others \cite{b1,b3}. In order for a machine learning model to recognize objects, it must be given a vast amount of data in the form of images containing the selected object(s) that the model is supposed to detect \cite{b4} specific objects within an image that it has never seen before \cite{b3,b5}.

Object detection of street-level objects is one of the use cases of object detection. It is mainly used in self-driving cars, UAVs, and surveillance cameras. By detecting these objects, it is possible to reduce accidents such as car crashes, collisions, and many more. This also allows authorities to monitor and track possible violations regarding road safety and traffic laws. Therefore, due to its high application value, it is chosen to be the basis of this research.

In the process of creating a machine learning model, other works typically utilize various algorithms and architectures such as  Convolutional Neural Networks (CNN) \cite{b6,b7}, Support Vector Machines (SVM) \cite{b4,b8,b9}, feature selection \cite{b10} and many more. The results of these various algorithms and architectures can also vary from one another, making each one unique and having its own best use case(s). As for object detection itself, some of the commonly used architectures and algorithms utilized in the past few years consist of Single Shot Multibox Detectors (SSD) \cite{b11}, You Only Look Once (YOLO) \cite{b12} and also Region Based Convolutional Neural Networks (RCNN) and Faster-RCNN \cite{b13,b14,b15}.

In the last few years, many studies have compared the performance and efficiency of object detection algorithms in a particular case and environment. However, research has not caught up enough with new object detection algorithm such as, YOLOv5. Currently, there is very little research related to YOLOv5 with its predecessors or other object detection algorithms, especially for detecting multiple objects on the road.

Several papers have compared object detection algorithms that are specific to one type of object on the road. Paper by S. Choyal and A. K. Singh \cite{b16} performed a comparison between Faster-RCNN and SSD MobileNet on traffic signs. The results found that Faster-RCNN is more accurate but requires more training time than MobileNet SSDs. Furthermore, a similar study was conducted by M. Shahud et al. \cite{b17} compared the subtypes of YOLOv3 on traffic signs and found that YOLOv3-tiny is 13\% less accurate than YOLOv3, but YOLOv3-tiny is very useful in real-time applications due to its high frame rate of up to 200 Frames Per Second (FPS).

One of the essential objects to detect on the road is humans. C.E.Kim et al. \cite{b18} performed comparisons on many object detection algorithms,  Faster-RCNN, YOLO, SSD, and R-FCN algorithms were compared. It was found that YOLOv3 is the most ideal for detecting people because it gives relatively accurate results in a reasonable time. Q.-C. Mao et al. \cite{b19} proposed an improvement to YOLOv3 with a Spatial Pyramid Pooling (SPP) layer so that YOLOv3 works better for detecting vehicles. His research succeeded in increasing the accuracy and error rate of YOLOv3 in vehicle detection.

Another object that often appears on the road, namely potholes, has been raised as a topic of object detection by P. Ping et al. \cite{b20}. Her paper found that YOLOv3 is the most efficient algorithm because its speed and detection results are more reliable than HOG, SSD, and Faster-RCNN. In addition, other objects such as trees also often appear on the roadside. A comparison of YOLOv3-SPP and YOLOv3 was carried out by Z. Yinghua et al. \cite{b21}. The result is that YOLOv3-SPP is claimed to be the most suitable for detecting trees.

From these studies, YOLOv3 has been chosen as the ideal object detection algorithm or the main research topic. Not only to detect objects on the road but also to detect other objects, such as mask detection, carried out by R. Liu and Z. Ren \cite{b22}. Where YOLOv3 can achieve good performance and lower inference time than Faster-RCNN.

Not many studies have been done on the new generation of YOLO to detect objects on the road. When there is, the comparison results are not as satisfactory, such as research by C. Kumar B. et al. \cite{b23} on the detection of multiple objects on surveillance cameras. Because the explanation of the experimental method is incomplete, and the metrics used in the evaluation are not standardized. However, in detecting other objects, the new generation of YOLO has been compared well, such as comparing YOLOv4 and YOLOv5 in the detection of insulators by E. U. Rahman et al. \cite{b24}. The result is surprising because YOLOv4 is better than YOLOv5.

There have been many studies conducted on the comparison of YOLO object detection. It can be said that the comparative study of object detection that often appears in road cases is on YOLOv3, Faster-RCNN, and SSD MobileNet. Although few studies are using the YOLOv5 algorithm, few have focused on comparing the YOLOv5 algorithm with its predecessors, explicitly detecting multiple objects on the road.

The following are the primary contributions of this paper after a thorough review of several object detection algorithms:
\begin{itemize}
\item An in-depth elaboration and analysis of object detection algorithms; SSD MobileNetv2, YOLOv3, YOLOv4, and YOLOv5 models.
\item A detailed evaluation of inference time, precision, recall, F1-score and mean Average Precision (mAP(.50)) of each model.
\end{itemize}

To present the tools and process in achieving the final results, the paper is arranged as such: The upcoming section presents materials and methods, whereas the next section describes the experimental setup and results in comparing the various object detection algorithms. Finally, the results and discussions of the experiment, conclusion and future works of this project are stated in more detail.

\section{Method}
\label{sec:method}
This section elaborates in more detail about the object detection algorithms: SSD and YOLO individually. Then followed by an elaboration of the dataset and dataset split, then the pre-processing and augmentation stage for real-time images in the dataset. Finally, each object detection algorithm is trained and compared. 

\subsection{One Stage Detectors} One-stage detectors refers to a deep neural network-based technique for object detection in real-world applications \cite{b25}. Where many modern detectors work in either one or two-stage manners \cite{b26}. Without the use of region proposals, one stage detectors are capable of detecting objects from all candidates. This results in a more compact one-stage detectors that are better suited for mobile devices \cite{b27}.

YOLO models and SSD MobileNetv2 are both one-stage detectors \cite{b27}. YOLO was developed with the goal of being able to directly predict classification scores and bounding boxes, without having any additional stages in generating region proposals \cite{b13}. Where SSD introduces anchors with multi-scale predictions that come from multi-layer convolutional features and Focal loss \cite{b28} in dealing with the problem regarding class imbalance that some one-stage detectors like RetinaNet faced \cite{b25}.

\subsection{SSD MobileNetv2} The SSD MobileNetv2 is an object detection model that computes an object’s bounding box and category from an input image using MobileNetv2 as a feature extractor \cite{b29}. The SSD is a feed-forward convolutional network object detection model that generates a set of fixed-size bounding boxes and scores for the object class instance(s) availability within those boxes \cite{b28}. Multi-reference and multi-resolution techniques are used by SSD. A set of anchor boxes of various sizes and aspect ratios are defined on different locations within an image using the multi-reference technique. Followed by the prediction of detection boxes which are done using SSD \cite{b30}. Finally, the process of object detection on several scales on different layers within the network is done using the multi-resolution technique \cite{b30}.

SSD has to follow up to two stages: extraction of the feature maps and then followed by the process of using convolutional filters to detect objects \cite{b29,b30,b31}. In extracting features, the SSD is modified with MobileNetv2 as feature extractor which uses an inverted residual structure and linear bottleneck to create more efficient layer structures \cite{b29, b32}. Compared to other real-time detectors such as SSD300, SSD512, YOLOv2, and MobileNetv1 on the COCO dataset, SSD MobileNetv2 can produce decent accuracy that can compete with other real-time detectors because the accuracy results are only slightly different while the required parameters are fewer \cite{b29}.

The MobileNetv2 itself is based on a bottleneck depth-separable convolution with residuals. The initial fully convolution layer comprises 32 filters and is preceded by 19 residual bottleneck layers, with ReLU6 serving as the quasi for increased low-precision computation \cite{b29}. 

\begin{figure}[!h]
    \centering
    \includegraphics[width=130mm]{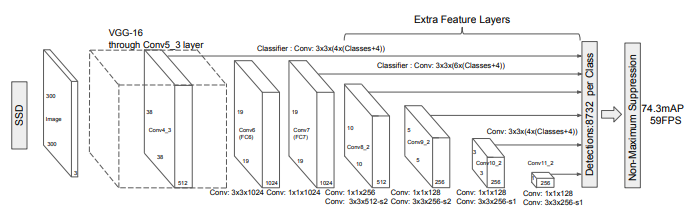}
    \caption{Layer architecture of a Single Shot Multibox Detector (SSD) network. Figure taken from~\cite{b28}}
    \label{fig:fig1}
\end{figure}

\subsection{YOLO Models}

Redmon et al. proposed the YOLO model in 2016 \cite{b12}, which is targeted for real-time processing. The main idea of YOLO was to apply a unique neural network to the whole image, then divide it into smaller regions and predict the probabilities and bounding boxes of each region in unison \cite{b30}. YOLO also utilizes a one-stage deep learning algorithm that implements convolutional neural networks in object detection \cite{b33}. Unlike other deep learning algorithms, which are incapable of detecting an object in a single run \cite{b34}, YOLO utilizes a single forward propagation across a neural network, making it more suitable for real-time applications \cite{b33}.

In the past years, YOLO has gone through several improvements \cite{b33}, which produces a number of versions such as YOLOv3, YOLOv4, YOLOv5 and many more. During the period during  which this paper was written, the latest version of YOLO is currently at YOLOv5. The main differences between the various YOLO versions are as follows. YOLOv3 utilizes a different feature extractor from the previous version. It uses respectively 3 × 3 and 1 × 1 convolutional network called Darknet-53 as the backbone feature extractor \cite{b35}, whereas YOLOv4 and YOLOv5 utilize the CSPDarknet53 as its backbone feature extractor. Besides that, the YOLOv3 and YOLOv4 uses the binary cross-entropy loss for class predictions \cite{b33, b35, b36}, whereas YOLOv5 also uses a binary cross entropy but also using the logits loss function as its loss function \cite{b33, b36}. The following Figure 2, 3 and 4 shows the layer structure within corresponding YOLO versions.

\begin{figure}[!h]
    \centering
    \includegraphics[width=100mm]{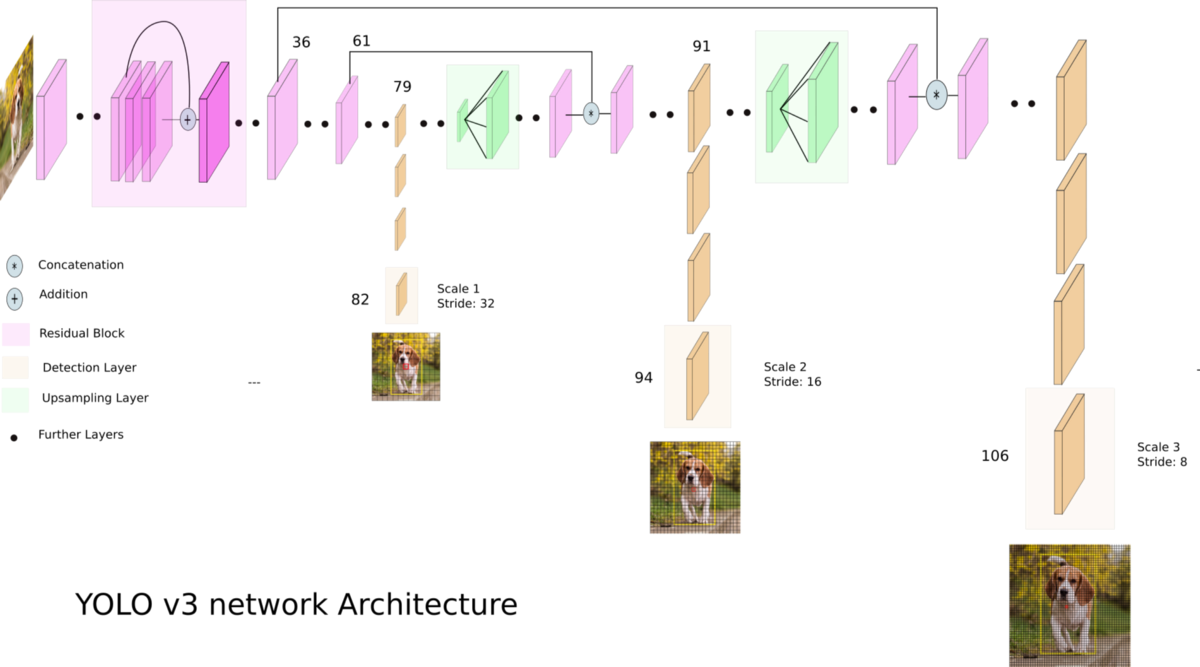}
    \caption{Layer Architecture of You Only Look Once (YOLO)v3. Figure taken from~\cite{b37}}
    \label{fig:fig2}
\end{figure}

YOLOv3 extracts features at three different scales where a number of convolutional network layers are added to the base feature extractor. In the last convolutional layer, it predicts the 3-D encoding box, objectness, and class predictions \cite{b35}.

\begin{figure}[!h]
    \centering
    \includegraphics[width=100mm]{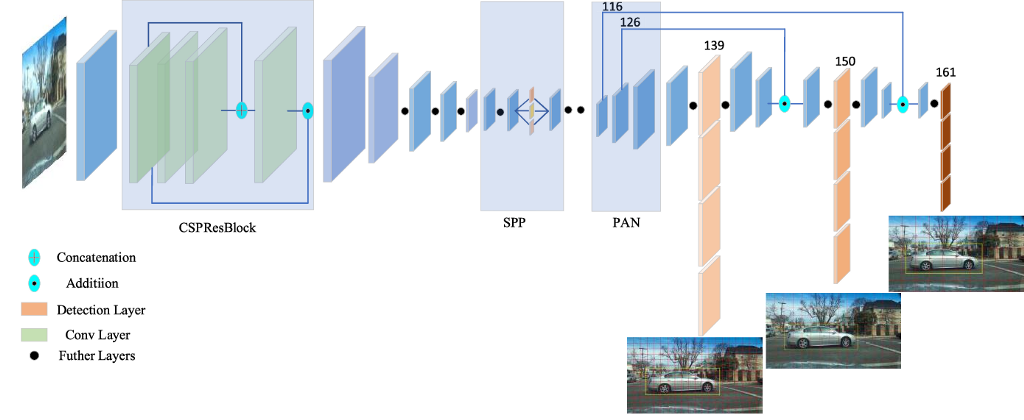}
    \caption{Layer Architecture of You Only Look Once (YOLO)v4. Figure taken from~\cite{b38}}
    \label{fig:fig3}
\end{figure}

\begin{figure}[!h]
    \centering
    \includegraphics[width=100mm]{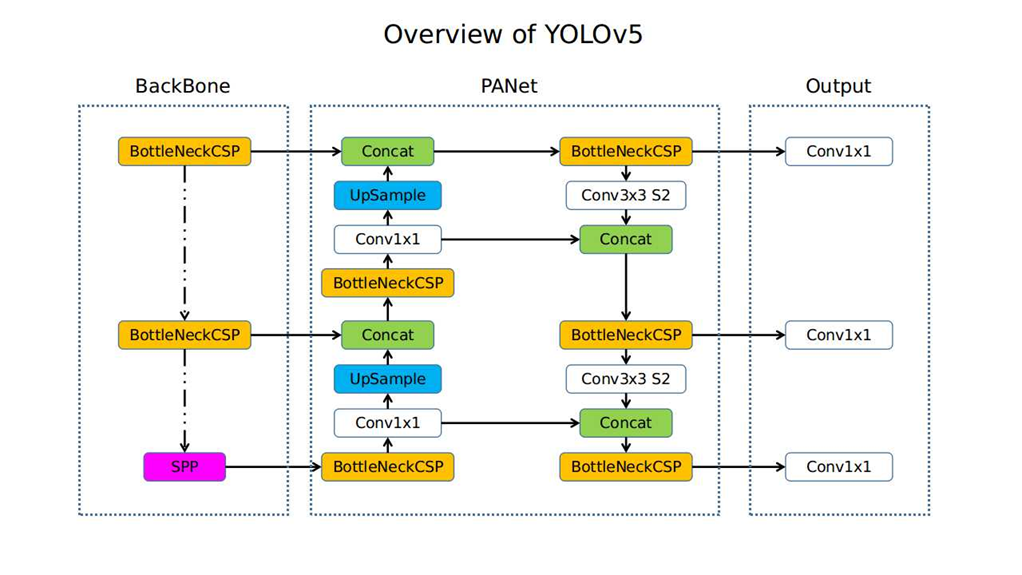}
    \caption{Layer Architecture of You Only Look Once (YOLO)v5. Figure taken from~\cite{b33}}
    \label{fig:fig4}
\end{figure}

The backbone feature extractor used in YOLOv4 and YOLOv5, CSPDarknet53, uses Darknet-53 with CSPNet \cite{b39} strategy, which merges two parts from feature maps partition of base layer using cross-stage hierarchy \cite{b40}. YOLOv4 utilizes CSPDarknet53, requiring 27.6 M parameters and containing 29 convolutional layers 3 × 3, Spatial Pyramid Pooling (SPP) and Path Aggregation Network (PAN) as the neck and YOLOv3 as the head of the network of YOLOv4. Furthermore, DropBlock is used as a regularization method for the backbone \cite{b40}.

The structure of YOLOv5 is similar to YOLOv4, where CSPDarknet53 is used for the backbone and to collect feature maps; the neck layers also use SPP and PAN \cite{b36}.  YOLOv5 was released in various such as YOLOv5s, YOLOv5m, YOLOv5l, and YOLOv5x. Where the difference between the model is the scaling multipliers width and depth of the network \cite{b36}.

\begin{figure}[!h]
    \centering
    \includegraphics[width=90mm]{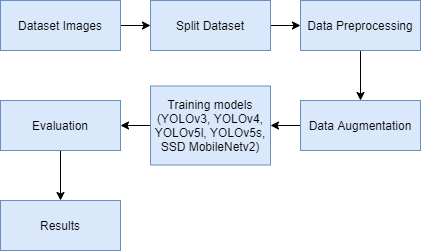}
    \caption{Experiment flow}
    \label{fig:fig5}
\end{figure}

\subsection{Dataset}
The dataset we use in this paper is the modified Udacity Self Driving Car Dataset that contains 3,169 images with 24,102 annotations. The classes and each of their balances contained are as the following in Table 1.

\begin{table}[!h]
\caption{Class Balance}
\label{table}
\setlength{\tabcolsep}{5pt}
\renewcommand{\arraystretch}{1.3}
\centering
\begin{tabular}{|p{100pt}|p{80pt}|}
\hline
Class& 
Amount 
\\
\hline
car& 
16,446 \\
trafficLight& 
4,790 \\
pedestrian& 
1,756 \\
truck& 
761 \\
biker& 
349 \\
\hline
\end{tabular}
\label{tab1}
\end{table}

\subsection{Split Dataset}
For the purpose of comparing various object detection algorithms, the experiment is conducted on the previously mentioned dataset. The dataset is split into three sets of data: Train, validation, and test data, which can be represented as 63\%, 18\%, 18\% of the data respectively.

\begin{table}[!h]
\caption{Dataset Distribution}
\label{table}
\setlength{\tabcolsep}{5pt}
\renewcommand{\arraystretch}{1.3}
\centering
\begin{tabular}{|p{100pt}|p{80pt}|}
\hline
Class& 
Amount 
\\
\hline
Train data& 
2,010 \\
Validation data& 
586 \\
Test data& 
573 \\
\hline
\end{tabular}
\label{tab2}
\end{table}

\subsection{Image Pre-Processing and Data Augmentation}
For the SSD MobileNetv2 and YOLO models, the image pre-processing consists of rescaling and cropping the image within a stated aspect ratio using the sizes of respective input layers. Following the rescaling process, data augmentation was applied to the dataset. Where it enables various geometric and color transformations to be made to the images, including scaling, transformation, and even color transformation. The data is preprocessed by rescaling each image into a size of 416 × 260 pixels while maintaining the aspect ratio of 16:10. Data augmentation such as hue shifting (between -25° to 25°). Noise (up to 5\% of pixels) and cutout (3 boxes with 10\% size each) are applied to the dataset images during the training.

\subsection{Parametrization of Object Detection Algorithms and Training Details}
In object detection, most models need to be trained using a set of input images and their associated ground truth boxes for each of their classes. In our approach, all of our models used, which are the SSD MobileNetv2 FPN-lite 320x320, YOLOv3, YOLOv4, YOLOv5l and YOLOv5s, are pre-trained with the COCO dataset before moving on to the phase which will be trained with the utilized dataset which will be elaborated in the next section.

In this paper, we compare the various object detection algorithms using real-time images, the main aspects that we compare each of the models consist of their precision, recall, F1-score, mAP values and also the inference time. In this experiment YOLOv3, YOLOv4, YOLOv5s and YOLOv5l use pytorch framework and SSD MobileNetv2 FPN-lite uses tensorflow framework. Since different frameworks required different iterations, YOLOv3, YOLOv4, YOLOv5s and YOLOv5l were trained for 100 epochs with Stochastic Gradient Descent (SGD) optimizer, and SSD MobileNetv2 FPN-lite was trained for 32000 steps with momentum optimizer.

In the training phase, the following table below shows hyperparameter settings that were used for each algorithm.

\begin{table}[!h]
\caption{Hyperparameter Settings}
\label{table}
\setlength{\tabcolsep}{5pt}
\renewcommand{\arraystretch}{1.3}
\centering
\begin{tabular}{|p{100pt}|p{80pt}|}
\hline
\multicolumn{2}{|c|}{Hyperparameter} \\
\hline
Batch Size& 
64 \\
Optimizer Momentum& 
0.9 \\
Weight Decay& 
0.001 \\
Learning Rate Scheduler& 
Cosine \\
Base Learning Rate& 
0.1 \\
Loss Gamma& 
2 \\
\hline
\end{tabular}
\label{tab3}
\end{table}

\subsection{Performance Metrics}
Several performance metrics were used to evaluate each algorithm. Precision, recall, F1-score, mAP, and inference time were chosen in this research. Performance metrics were evaluated on a test dataset.

\begin{equation}
Precision= {\frac {TruePositive}{TruePositive+FalsePositive}}
\end{equation}
\begin{equation}
Recall= {\frac {TruePositive}{TruePositive+FalseNegative}}
\end{equation}
\begin{equation}
F1-Score= {2 \cdot \frac {Precision \cdot Recall}{Precision+Recall}}
\end{equation}
\begin{equation}
mAP= {\frac {\sum _{q=1}^{Q} AveP(q)}{Q}}
\end{equation}

Equation (1), Precision finds the percentage of correct predictions over false positive and true positive, which measure how accurate the prediction results are. Equation (2), Recall defines how well the algorithm finds all the positive cases. Equation (3), F1-Score is the balance between precision and recall, the value ranging between 0 to 1. Higher F1-Score means a better balance on precision and recall.

Equation (4), mean average precision of all classes is used to determine mAP. mAP is considered to be the measure to analyze the overall performance of an object detection algorithm. In this experiment, different Intersection over Union (IoU) values were used for mAP. The percentage match area of intersection between the predicted box and the ground-truth box is measured by IoU. mAP@.5 means the threshold of IoU for mean average precision is 0.5.

Furthermore, this experiment also uses mAP@.5:95, which is the average mAP over several IoU thresholds ranging from 0.5 to 0.95. Inference time shows how fast the algorithm predicts in real-time; the less time it takes to predict, the better it is for a real-time application.

The experiment was evaluated using the computer specification in table 4.

\begin{table}[!h]
\caption{Computer Specification}
\label{table}
\setlength{\tabcolsep}{5pt}
\renewcommand{\arraystretch}{1.3}
\centering
\begin{tabular}{|p{100pt}|p{150pt}|}
\hline
\multicolumn{2}{|c|}{Specification} \\
\hline
Central Processing Unit (CPU)& 
8th Gen. Intel® Core™ i5-8600K Processor 3.60GHz\\
Graphics Processing Unit (GPU)& 
NVIDIA® GeForce® GTX 1050 TI \\
Random-access Memory (RAM)& 
16GB DDR4 3200MHz \\
Storage& 
500GB SSD up to 560/530 MB/s \\
Operating System& 
Windows 10 21H1 \\
\hline
\end{tabular}
\label{tab4}
\end{table}

\section{Results and Discussion}
Table 5 conveys the testing results based on the performance metric of each object detection algorithm from the previous experiments conducted.

\begin{table}[!h]
\centering
\caption{Quantitative Testing Results of Various Object Detection Algorithms}
\label{table}
\setlength{\arrayrulewidth}{0.1mm}
\setlength{\tabcolsep}{5pt}
\renewcommand{\arraystretch}{2.1}
\resizebox{.80\textwidth}{!}{
\begin{tabular}{|c|c|c|c|c|c|}
\hline
Measure & SSD MobileNetv2 FPN-Lite & YOLOv3 & YOLOv4 & YOLOv5l & YOLOv5s \\
\hline
\setlength{\tabcolsep}{1pt}
Precision & 0.354 & 0.767 & 0.569 & 0.780 & 0.670 \\

Recall & 0.228 & 0.535 & 0.589 & 0.545 & 0.501 \\

F1-Score & 0.277 & 0.630 & 0.579 & 0.642 & 0.573 \\

mAP@.5 & 0.315 & 0.583 & 0.582 & 0.593 & 0.530 \\

mAP@.5:95 & 0.152 & 0.313 & 0.304 & 0.313 & 0.260 \\

Inference Time (ms) & 6.30 & 27.60 & 27.90 & 25.40 & 8.50 \\
\hline
\end{tabular}}
\label{tab5}
\end{table}

\begin{figure}[!h]
    \centering
    \includegraphics[width=165mm]{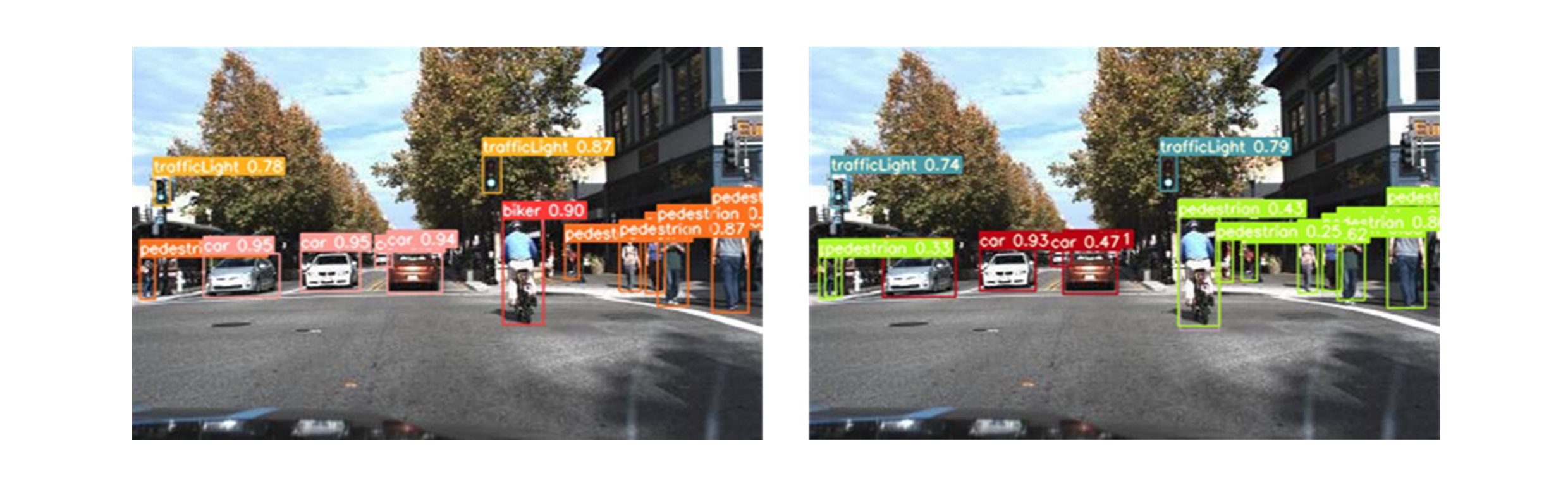}
    \caption{Detection result illustrated on sample test image. Left is YOLOv3 and right is YOLOv4.}
    \label{fig:fig6}
\end{figure}

\begin{figure}[!h]
    \centering
    \includegraphics[width=165mm]{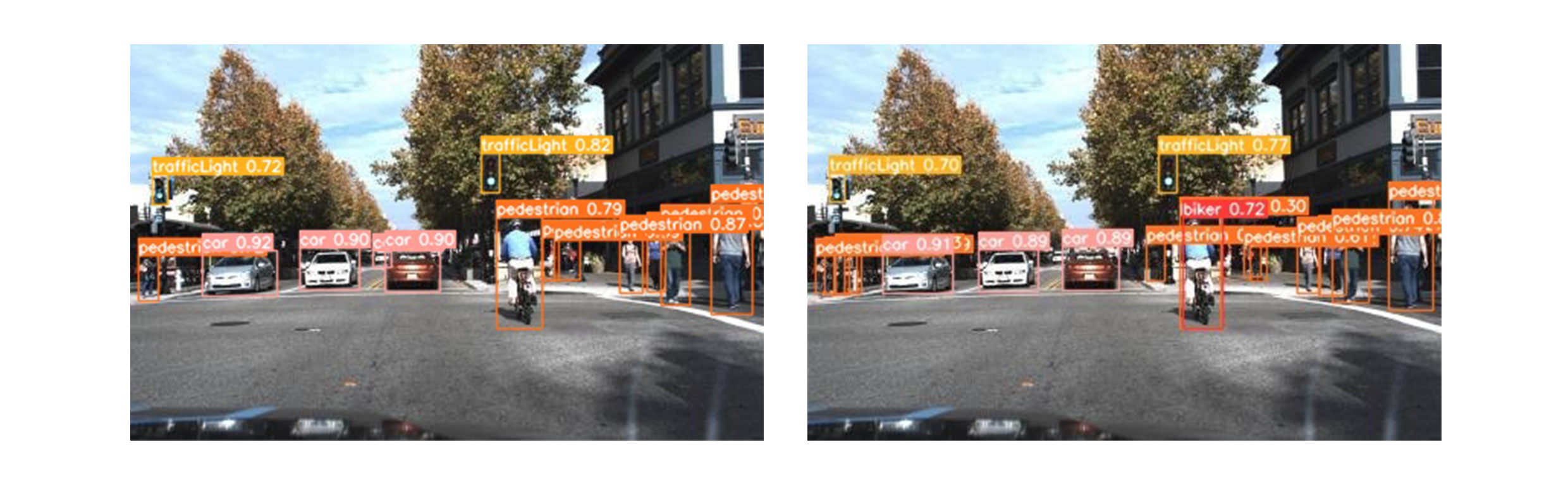}
    \caption{Detection result illustrated on sample test image. Left is YOLOv5l and right is YOLOv5s.}
    \label{fig:fig7}
\end{figure}

\begin{figure}[!h]
    \centering
    \includegraphics[width=80mm]{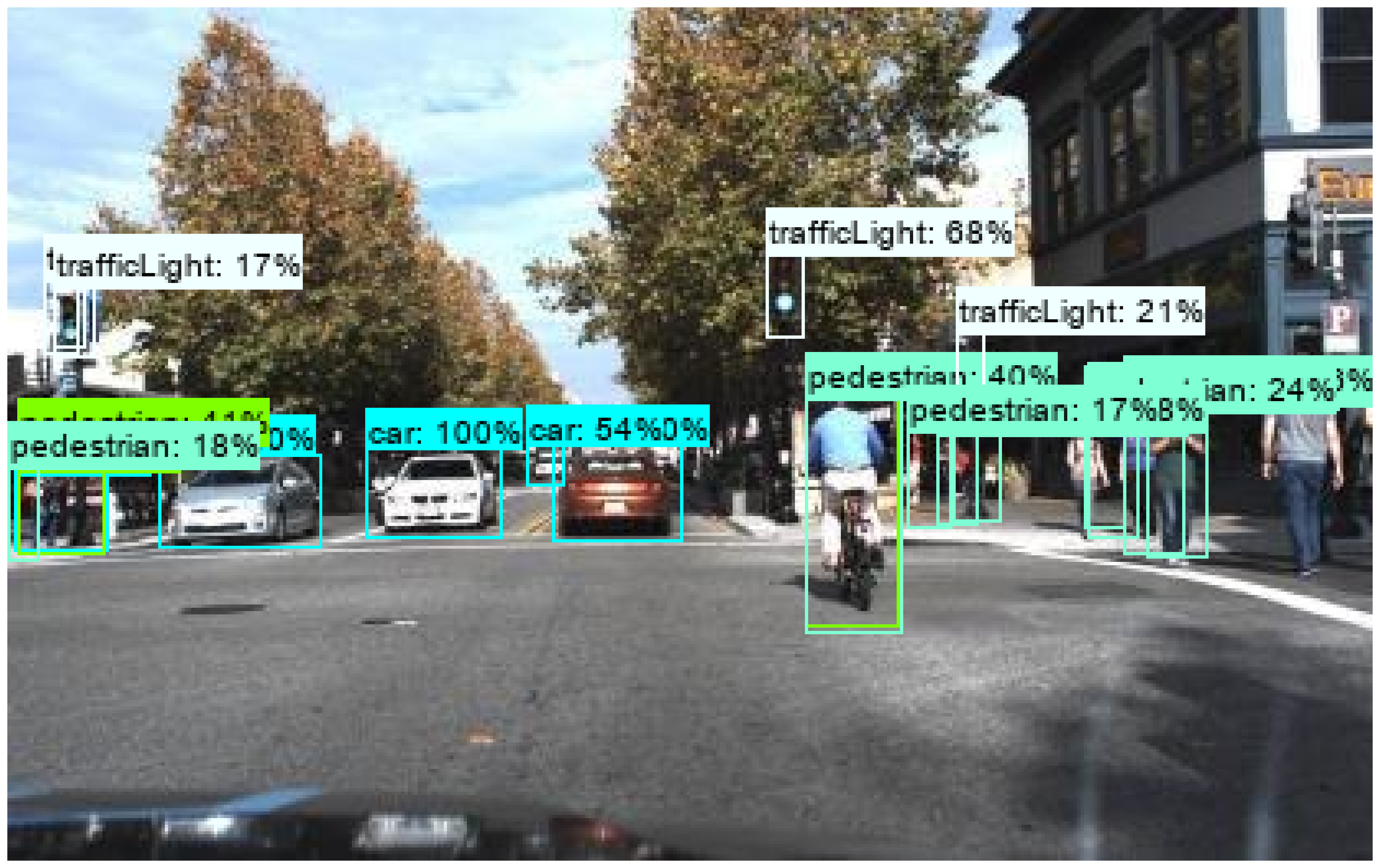}
    \caption{SSD MobileNetv2 detection result illustrated on sample test image.}
    \label{fig:fig8}
\end{figure}

YOLOv5l presents the highest F1-Score, mAP@.5, and mAP@.5:95 compared to other algorithms that were evaluated. This means that YOLOv5l is the most accurate algorithm in this experiment, but compared to the previous YOLO algorithm, the difference in the mAP results was not significant. It is also shown that YOLOv4 strangely performed worse than YOLOv5l and YOLOv3 in terms of F1-Score, precision, and mAP by a little. YOLOv4 model’s inference time is also the slowest compared to all other models in this experiment with 27.9ms. It is also seen that YOLOv5l surpasses both YOLOv4 and YOLOv3 in terms of mAP@.5 and inference time.

\begin{figure}[!h]
    \centering
    \includegraphics[width=120mm]{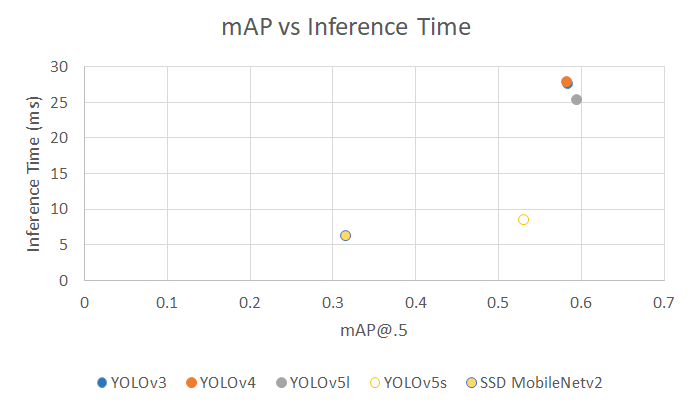}
    \caption{mAP vs Inference Time of  Various Object Detection Algorithms.}
    \label{fig:fig9}
\end{figure}

Compared to all object detection algorithms that were evaluated in this experiment, SSD MobileNetv2 FPNLite performed the worst in accuracy with mAP@.5 of only 0.315, but it is the fastest algorithm in this experiment with only 6.3ms of inference time. The second fastest object detection algorithm is YOLOv5s, with 8.50ms F1-Score and 0.530 mAP@.5, which is only 11\% worse, and the mAP@.5:95 is 17\% worse than the most accurate algorithm in this experiment, YOLOv5l.

\subsection{Conclusion}
This paper compared various object detection algorithms which can detect street-level objects. Five algorithms, including  SSD MobileNetv2 FPN-lite 320x320, YOLOv3, YOLOv4, YOLOv5l and YOLOv5s, were trained and evaluated on a modified Udacity Self Driving Car Dataset, the dataset is also augmented by rescaling, hue shifting the image, as well as adding noise to the image. Results found that YOLOv5l is the most accurate algorithm compared to the others and SSD MobileNetv2 FPN-lite is among the fastest for detecting street-level objects. However, further analysis indicates that YOLOv5s is the ideal algorithm for real-time application such as self-driving cars in detecting street-level objects because it provides relatively accurate results in a reasonable time. Future work may include future object detection models and comparing their results to existing models. Also by using a larger dataset for the training, validating and testing processes.

\bibliographystyle{unsrt}  
\bibliography{references}

\end{document}